\documentclass[letterpaper]{article} 
\usepackage{aaai23}  
\usepackage{times}  
\usepackage{helvet}  
\usepackage{courier}  
\usepackage[hyphens]{url}  
\usepackage{graphicx} 
\usepackage{natbib}  
\usepackage{caption} 
\usepackage{algorithm}
\usepackage{algorithmic}

\usepackage{amsmath}
\usepackage{amsfonts}

\usepackage[title]{appendix}

\usepackage{wrapfig}
\usepackage{color}

\usepackage{listings}

\usepackage{newfloat}
\usepackage{listings}
\DeclareCaptionStyle{ruled}{labelfont=normalfont,labelsep=colon,strut=off} 
\floatstyle{ruled}
\newfloat{listing}{tb}{lst}{}
\floatname{listing}{Listing}
\title{ Long $N$-step Surrogate Stage Reward to Reduce Variances of Deep Reinforcement Learning in Complex Problems}

\author{
    Junmin Zhong\textsuperscript{\rm 1}, 
    Ruofan Wu\textsuperscript{\rm 1}, 
    Jennie Si\textsuperscript{\rm 1}\thanks{Corresponding author: si@asu.edu}
}
\affiliations{
    \textsuperscript{\rm 1}Arizona State University\\

    Tempe, Arizona 85287 USA\\

}

\usepackage{bibentry}

\begin{document}

\maketitle

\begin{abstract}
High variances in reinforcement learning have shown impeding successful convergence and hurting task performance. As reward signal plays an important role in learning behavior, multi-step methods have been considered to mitigate the problem, and are believed to be more effective than single step methods.
However, there is a lack of comprehensive and systematic study on this important aspect  to demonstrate the  effectiveness of multi-step methods in solving highly complex continuous control problems.
In this study, we introduce a new  long $N$-step surrogate stage (LNSS) reward approach to effectively account for complex environment dynamics while previous methods are usually feasible for limited numbers of steps. The LNSS method is simple,  low computational cost, and applicable to value based or policy gradient reinforcement learning. 
We systematically evaluate LNSS in OpenAI Gym and DeepMind Control Suite  to address some  complex benchmark environments that have been challenging  to obtain good results by DRL in general. We demonstrate performance improvement in terms of total reward, convergence speed, and coefficient of variation (CV)  by LNSS. We also provide analytical insights on how LNSS exponentially reduces the upper bound on the variances of $Q$ value  from a respective single step method.

\end{abstract}

\section{Introduction}
Great progress has been made in deep reinforcement learning (DRL) aiming at solving highly complex and continuous control problems involving high-dimensional inputs and large action spaces. Several algorithms, such as DQN \cite{mnih2013playing}, deep deterministic policy gradient (DDPG) \cite{lillicrap2015continuous}, proximal policy optimization (PPO) \cite{schulman2017proximal},
Soft actor critic (SAC) \cite{haarnoja2018soft} and twin delayed DDPG (TD3) \cite{fujimoto2018addressing}, have demonstrated their potential in tasks such as   hopper and  walker2d in OpenAI Gym (GYM) and Deepmind Control Suite (DMC). However, current solution approaches still face challenges to solve  very complex tasks such as humanoid to achieve good results \cite{duan2016benchmarking,pardo2020tonic}.
 The high variance problem intrinsic to the trial and error nature of reinforcement learning has been considered one of the major road blocks as  it causes low data efficiency, slow learning or  even instability of learning  and poor task performance \cite{henderson2018deep,duan2016benchmarking}. A sign a variance problem is  significant oscillation during late stages of learning.


Reducing learning variance has been approached from different perspective. Some methods aim at reducing policy variance. The TD3 algorithm \cite{fujimoto2018addressing} uses double $Q$-learning to clip the target value in order to reduce  overestimation errors. With the inter-play  between the actor and the critic, limiting estimation errors helps reduce variance in the policy.
PPO \cite{schulman2017proximal} uses the trust region method and a clipped objective to help bound  policy updates and thus reduce the variance of learned policy. SAC \cite{haarnoja2018soft} uses a maximum entropy idea to maximize both the expected return and the expected entropy of a policy. D4PG \cite{barth2018distributed}, instead of learning a value function, learns a distribution over returns  to help mitigate $Q$-value overestimation errors, an idea similar to TD3.  

Averaging value functions over different iterations is another idea to benefit from data used in different iterations. Averaged DQN \cite{anschel2017averaged} has been shown helpful in reducing overestimation errors and stabilizing learning. As shown in MMDDPG \cite{meng2021effect}, by averaging over a set of target Q values with different step lengths, learning stability has improved in DDPG.

Eligibility trace in TD($\lambda$) uses information of many multi-step target values to create a compound target and it has been shown to outperform either  TD methods or Monte Carlo methods \cite{sutton2018reinforcement}. 
The tree back up  \cite{precup2000eligibility}  corrects target discrepancy by modulating each step with a target policy probability. And the $Q(\sigma$) \cite{sutton2018reinforcement} unifies and generalizes the existing multi-step TD methods by the hyper parameter $\sigma$ to  allow a mixture of sampling and expectation approaches. However, these methods are to be demonstrated effective on complex continuous control problem.

Rollout methods \cite{bertsekas2010rollout,tesauro1994td,silver2018general} 
have enabled the success of some of the most important milestone cases  such as Backgammon \cite{tesauro1994td} and Alpha-zero \cite{silver2018general} in machine learning history. Although rollout has been demonstrated in discrete state and control problems, variations of rollout are  believed to be effective also in continuous control tasks.

In this paper, we devise a new multi-step method, namely the long $N$-step surrogate stage (LNSS) reward method, aiming to provide variance reduction for DRL in complex continuous control problems.

\section{Related Work}

The $n$-step methods evolve around the idea of utilizing trajectories with length $n$ to update the target function. 
Many modern DRL algorithms such as A2C \cite{mnih2016asynchronous} and D4PG \cite{barth2018distributed} use $n$-step methods to update target value, and they have been shown  effective in continuous control tasks. PPO \cite{schulman2017proximal} uses a truncated version of generalized advantage estimation (GAE) to update policy. Rainbow \cite{hessel2018rainbow} integrates $n$-step learning on top of DQN algorithm and results in a speed up in learning and an improvement of the final performance on Atari 2600 games. Mean reward method in multi-step Q-learning \cite{yuan2019novel} is another way to use $n$-step trajectories by simply  using a mean reward to perform a single step update.

Reward estimation \cite{van2013efficient,feinberg2018model,silver2017predictron}, instead of directly using an actual $n$-step reward,  uses multi-step imaginary rollouts as predicted reward for planning rather than training a value function. The $\lambda$-prediction \cite{silver2017predictron} requires an extra network.
Model-based value expansion \cite{feinberg2018model} uses the imaginary reward from a model to estimate the value of state value functions.

Using longer trajectories with a larger $n$ 
is expected to  improve learning performance as shown theoretically \cite{hernandez2019understanding}. 
However to date, the $n$-step DRL methods have only been demonstrated by using relatively small $n$. 
Rainbow \cite{hessel2018rainbow} uses $n = 1, 3, 5$ to evaluate performance of 57 Atari 2600 games 
and reported $n = 3$ as the best performing trajectory length. D4PG \cite{barth2018distributed} uses $n = 1, 5$ to evaluate  a variety of continuous control tasks in DMC and results show that  $n = 5$ perform uniformly better than others.
For the reward estimation method, \cite{silver2017predictron} reported results based on $n=6$ steps using neural networks and  show that $n$-step significantly outperformed model-free algorithms in random maze problem. Although up to 16 steps have been demonstrated for estimated rewards,  the neural network prediction model has to  be exponentially larger as step numbers increase. In \cite{feinberg2018model}, $n=30$ steps was tested. But in their result, $n=$2, 10, 30 steps for  half-cheetah in GYM does not make significant difference. \cite{van2013efficient} showed results with $4$  and $15$ successor states in a discrete action maze task where more  successor states has better average return. However, $3$,$5$ or even $30$ steps still are relative small for  continuous benchmark tasks which usually have a maximum of 1000 steps in an episode. Therefore, there needs a method that can evaluate longer $n$-step trajectories.

In this paper, we propose a new idea, the LNSS method, to enable effective use of long trajectories for variance reduction especially for addressing complex problems. The LNSS method can be implemented on top of  any DRL algorithms. When selecting the DRL algorithm, we take the following into consideration to evaluate our LNSS method. 
Among all DMC based benchmark results \cite{pardo2020tonic}, D4PG \cite{barth2018distributed} outperforms other DRL algorithms such as DDPG \cite{lillicrap2015continuous}, TD3 \cite{haarnoja2018soft}, and PPO \cite{schulman2017proximal}.  SAC 
 \cite{haarnoja2018soft} outperforms other DRL algorithms in GYM. PPO and TD3 has similar performance that have relatively good performance in GYM but struggles to achieve good result in complex DMC tasks such as humanoid-walk, fish-swim. 
With the above benchmark results, we chose to systematically test the idea of LNSS on TD3 as our base algorithm for two reasons. First, there is a lack of multi-step methods associated with TD3 to test complex benchmark environments. Additionally, TD3 has been shown highly effective in benchmark studies on relative simple tasks such as Hopper and Walker2d, but it still faces challenge in addressing complex tasks such as humanoid and fish-swim in DMC \cite{duan2016benchmarking,pardo2020tonic}.  

{\textbf {Contributions}}. 1) We introduce a new, model-free long $N$-step surrogate stage (LNSS)  reward estimator for infinite horizon discounted reward DRL problems. 2) We provide analytical insight on how  LNSS with long $N$ steps exponentially reduces the upper bound on the variance of $Q$ value from a respective single step method. 3) We  present a systematic evaluation of LNSS in solving complex, high-dimensional control tasks in GYM and DMC, such as fish swim, quadruped walk, and humanoid walk, where most DRL methods have not been able to achieve good results. 




\section{Background}
{\textbf {Reinforcement Learning}}. In this work we consider a standard reinforcement learning setting where agents interacts with its environment in discrete time. At each time step $k$, the control agent observes a state $s_k \in \mathcal{S}$ and select an action $a_k \in \mathcal{A}$  based on its policy $\pi : \mathcal{S} \to \mathcal{A}$, namely, $a_k = \pi(s_k)$, and receives a scalar reward $r(s_k,a_k) \in \mathcal{R}$ (use $r_k$ as short hand notation). We consider the infinite horizon discounted  return 
of the form $R_k= \sum_{t=k}^{\infty} \gamma^{t-k}r_t$, where $0 < \gamma < 1$.
An optimal policy is one that maximizes the return $R_k$. Successes of RL algorithms rely on action value function to approximate the return:
\begin{equation}\label{Eq:apprx_return}
    Q(s_k,a_k) = \mathbb{E}[R_k|s_k,a_k],
\end{equation}
where the value function estimation is conditioned on a current state and action pair. 

{\textbf {DRL Algorithm}}. The following background is needed in our LNSS-based DRL solutions to solving complex GYM and DMC tasks. 



DDPG \cite{lillicrap2015continuous} is a well-established off-policy policy gradient method. As an  actor-critic (AC) algorithm, it contains two steps of policy evaluation (computing value function for a policy) and policy improvement (using value function to find a better policy) \cite{sutton2018reinforcement}. Policy ($\pi$) is called an actor and action value function ($Q(s_{k}, a_{k})$) is call a critic where both the actor and the critic are estimated by deep neural networks.  Most AC methods are based on Bellman equation. The target equation becomes $y = r_k + \gamma Q(s_{k+1}, a_{k+1})$ so that critic is updated by minimizing the loss function with respect to the weights  ($\theta$) as:
\begin{equation}\label{Eq.Loss}
     L\left(\theta\right) =\mathbb{E}_{s\sim p_{\pi}, a \sim \pi} [(y - Q(s_{k}, a_{k}))^2].
\end{equation}
The actor weights can be updated by taking gradient of the $Q$ value:
\begin{equation}\label{Eq.Actor_up}
\begin{aligned}
    &\nabla_{\theta} Q(s_{k}, a_{k}) \\
    &=\mathbb{E}_{s \sim p_{\pi}}\left[\left.\nabla_{a} Q^{\pi}(s_{k}, a_{k})\right|_{a_k=\pi(s_k)} \nabla_{\theta} \pi(s_k)\right].
\end{aligned}
\end{equation}

TD3 \cite{fujimoto2018addressing} is based on DDPG but uses a clipped double $Q$ network idea, where $Q_{\theta_j}(s_k,a_k)$ ($j= 1,2$)  represent the two $Q$ values. It takes the lesser value between the two $Q$ values, thus the target function $y$ becomes:
\begin{equation}
    y = r_k + \gamma \min_{j=1,2} Q_{\theta_j}(s_{k+1}, a_{k+1}).
\end{equation}
Results show that this twin delayed double $Q$ network approach has effectively limited overestimation error. As in DDPG,  each of the $Q$ values is updated by minimizing the loss function $ L\left(\theta_{j}\right)$ with respect to their weights:
\begin{equation}
     L\left(\theta_{j}\right) =\mathbb{E}_{s\sim p_{\pi}, a \sim \pi} [(y - Q_{\theta_j}(s_{k}, a_{k}))^2].
\end{equation}
The actor network is updated the same way as DDPG (Equation (\ref{Eq.Actor_up})).


As is well known, $r_k$ in the Bellman equation play an important role. It thus has attracted great attention for potentially imporoving learning and task performance. This paper is related to two such approaches, namely,  single step methods (such as DDPG, and TD3) and $n$-step methods (such as D4PG and those explored in this study). More details about these single and $n$-step methods can be found in Appendix \ref{appendix:background}.

{\textbf {Single Step Method}}.  In single step RL, the $Q$ value is updated by using TD error or Bellman error with a single step return as follows,
\begin{equation}\label{Eq:single_step}
Q\left(s_{k}, a_{k}\right) = r_{k} +\gamma Q\left(s_{k+1}, a_{k+1}\right).
\end{equation}
 The update rule is also known as backup operation since it transfers information from one step ahead back to the current state. 


{\textbf {$n$-step Method}}. Using $n$-step reward trajectory for faster reward propagation in RL has long been investigated  \cite{watkins1989learning,de2018multi,barth2018distributed,hessel2018rainbow}. In $n$-step methods, the value function $Q\left(s_{k}, a_{k}\right)$ is driven by an $n$-step return as
\begin{equation}\label{Eq:n_step}
    Q\left(s_{k}, a_{k}\right)  = \sum_{t=k}^{k+n-1} \gamma^{t-k} r_{t} +\gamma^{n} Q(s_{k+n}, a_{k+n}). 
\end{equation}
The $n$-step return is expected to help agents learn more efficiently by allowing the $n$-step return affecting multiple state action pairs within one update and gain information over the $n$ step trajectory interval. 

The mean reward method \cite{yuan2019novel} utilizes  $n$-step trajectory information based on the following mean reward, 
\begin{equation}\label{Eq:mean_reward}
    r_{avg} = \frac{1}{n}(\sum_{t=k}^{k+n-1} r_{t}), 
\end{equation}
which is used to update the Bellman equation as in a single step method, 
\begin{equation}
    Q(s_k,a_k) = r_{avg} + \gamma Q(s_{k+1}, a_{k+1}).
\end{equation}


\section{Long $N$-step Surrogate Stage (LNSS) Reward } \label{SECTION.LNSS}
\label{sec:LNSS}
In this section, we introduce LNSS based on infinite horizon discounted reward formulation of reinforcement learning.
Let $G(s_{k:k+N-1},a_{k:k+N-1}) \in \mathbf{R}$ (use $G_k$ as short hand notation) denote the discounted $N$-step reward, i.e.,
\begin{equation}
G_k = \sum_{t=k}^{k+N-1} \gamma^{t-k}  r_t,
\end{equation}
where $r_k$ is the $k$th stage reward. Let $r'$ denote a surrogate stage reward in place of $r_k$. It is estimated based on $N$-step trajectories. Similarly, we  define a surrogate discounted $N$-step reward $G'$ as 
\begin{equation}\label{Eq:G'}
\begin{aligned} 
    G' &= \sum_{t=k}^{k+N-1} \gamma^{t-k} r' =  \frac{\gamma^N - 1}{\gamma - 1}r'.\\
\end{aligned}
\end{equation}
We then propose the surrogate stage reward $r'$ to be
\begin{equation}\label{Eq:r'}
\begin{aligned} 
    r' &= G_k*\frac{\gamma - 1}{\gamma^N - 1}. 
\end{aligned}
\end{equation}

This surrogate stage reward $r'$ as formulated in Equation (\ref{Eq:r'}) relies on a discounted reward of an $N$-step trajectory,  from time step $k$ to step $(k+N-1)$, from the 
stored experiences in a temporary replay buffer $\mathbb{D}'$. For a training episode of $T$ steps  $[0,1,2...,T]$, the $\mathbb{D}'$ is a moving window of size $N$ from the initial state $s_0$ until the terminal state $s_T$.  


In implementations of LNSS, there are cases when the experience buffer has less than $N$ samples to form an $N$-step trajectory. Consider a task episode of length $T$. The surrogate reward $r'$ in LNSS is determined according to the relative position of current stage $k$  in relation to $N$ and $T$.

1) If $k + N - 1 \leq T$, which signifies a sufficient number of experience samples  to compute $r'$ with full $N$ steps, then
\begin{equation}\label{Eq:calc_r'1}
    r' = (\sum_{t=k}^{k+N-1} \gamma^{t-k} r_t)*\frac{\gamma - 1}{\gamma^N - 1}.
\end{equation}

2) If $k + N - 1 > T$, which indicates a shortage of experience samples to compute $r'$ with full $N$ steps, then LNSS computes $r'$ based on what is available as shown in Equation (\ref{Eq:calc_r'2}) below,

\begin{equation}\label{Eq:calc_r'2}
    r' = (\sum_{t=k}^{T} \gamma^{t-k} r_t)*\frac{\gamma - 1}{\gamma^{T-k+1} - 1}.
\end{equation}

Once  $r'$ is obtained, $r'$ and state action pairs $(s_k,a_k,s_{k+1})$ will append as a new transition $(s_k,a_k,r',s_{k+1})$  stored into the memory buffer $\mathbb{D}$. 

Note that many DRL algorithms \cite{schulman2017proximal,hessel2018rainbow,barth2018distributed} use distributed learning procedure to accelerate experience sample collection. We use the same technique to speed up sampling experiences. Then  a DRL algorithm is ready to update the $Q$ value and the respective policy based on mini-batch data from the memory buffer. In general form, we have 

\begin{equation}\label{Eq:Q}
\begin{aligned}
    Q_{i+1}(s_k,a_k) &= r' + \gamma^n Q_i(s_{k+n}, \pi_i(s_{k+n})), \\
    \pi_i(s_k) &= arg \max_{a_k} Q_i(s_k,a_k), \\
\end{aligned}
\end{equation}
where $i$ is iteration number, and $n$  is the $n$-step updates for $Q$ function.
Putting the above two equations together, we have
\begin{equation}\label{eq:multistep}
    Q_{i+1}(s_k,a_k) = r' + \gamma^n \max_{a_{k+n}} Q_i(s_{k+n}, a_{k+n}).
\end{equation}

{\textbf {Remark 1}}. Note that, for most $n$-step methods governed by  Equation (\ref{Eq:n_step}), the same $n$ is used for accumulating reward over multiple steps and for updating the $Q$ function. However, in our LNSS, in Equation (\ref{Eq:r'}), $N$ is the number of steps for accumulating rewards and it  can be different from $n$ in Equation (\ref{Eq:Q}) for $Q$ value update. Our surrogate stage reward $r'$ approximates a stage reward $r_k$, thus using $n=1$ naturally helps keep Bellman equation balanced.
Our motivation of creating LNSS is to enable large number of step rewards (i.e., $N >> 1$) to be considered in updating $Q$ value and thus affecting the policy by several state-action pair transitions to capture complex task dynamics. Our formulation has given us the flexibility to address challenges associated with complex problems for the following additional considerations. 

As previously discussed,  $n$ in Equation (\ref{Eq:n_step}) is up to $5$ steps as a maximum in D4PG \cite{barth2018distributed} and $5$ steps in Rainbow \cite{hessel2018rainbow}. However the maximum step length of complex benchmark environment in Gym and DMC are usually $1000$ steps, a short trajectory of $5$ steps may not capture intricate dynamics of complex problems. Therefore LNSS provides the flexibility of using long $N$ steps such as $N = 50, 100$ steps in Equation (\ref{Eq:r'}) and small $n = 1$ in Equation (\ref{Eq:Q}). 




\section{Variance Analysis} 

We now analyze the behavior of a single step actor-critic RL and our LNSS actor-critic RL with $n=1$. We consider the infinite horizon discounted reward formulation of RL (with $0<\gamma<1$). 
Specifically, we show that the upper bound on the variance in $Q$ value due to LNSS differs by an exponential factor from that of a single step AC. 
As this upper bound reduces exponentially as $N$ increases, it suggests significant variance reduction by using LNSS from using single step reward. 



We first represent the Q values using single step reward $r_k$ in Equation (\ref{Eq:single_step}) and using surrogate reward $r'$ from LNSS in Equation (\ref{Eq:Q}), respectively as follows,
\begin{equation}\label{Eq:varQ}
\begin{aligned}
    var[Q_{i+1}(s_{k},a_{k})] &= var[r_k] + var[\gamma Q_i(s_{k+1},a_{k+1})] \\
    &+ 2 cov[r_k, \gamma Q_i(s_{k+1},a_{k+1})].
\end{aligned}
\end{equation}

\begin{equation}\label{Eq:varQQ}
\begin{aligned}
    var[\mathbb{Q}_{i+1}(s_{k},a_{k})] &= var[r'] + var[\gamma \mathbb{Q}_i(s_{k+1},a_{k+1})] \\
    &+ 2 cov[r', \gamma \mathbb{Q}_i(s_{k+1},a_{k+1})].
\end{aligned}
\end{equation}

{\textbf {Lemma 1}}. Assume $\{r_k\}$ is IID and drawn 
from the memory buffer $\mathbb{D}$.
Let  $Q_i(s_{k+1},a_{k+1})$ in Equation (\ref{Eq:Q}) be the $i$-th approximated return to solve Equation (\ref{Eq:apprx_return}). We then have the following,
\begin{equation}\label{Eq:covreq0}
    cov(r_k,r_{j \neq k}) = 0,
\end{equation}
\begin{equation}\label{Eq:covrQeq0}
    cov(r_k,Q_i(s_{k+1},a_{k+1})) = 0.
\end{equation}

{\textbf {Proof}}.
Given that  $\{r_k\}$ is IID, we reach Equation (\ref{Eq:covreq0}) immediately.
Based on Equation (\ref{Eq:apprx_return}), the conditional probability distribution of $Q_i(s_{k+1},a_{k+1})$ only depends on  state action pair $(s_{k+1},a_{k+1})$. It thus leads to  Equation (\ref{Eq:covrQeq0}).

Based on Lemma 1, we now analyze the variance  of the $Q$ value sequence.

{\textbf {Theorem 1}}. Consider the variances of two $Q$ value sequences, denoted as {${Q_i}$} and {$\mathbb{Q}_i$}, in Equation (\ref{Eq:varQ}) and Equation (\ref{Eq:varQQ}), which are obtained respectively from a single step method and an LNSS method. Additionally, assume that  $Q_0 = var[Q_0] = 0$ and $\mathbb{Q}_0 = var[\mathbb{Q}_0] = 0$. Let the IID reward $\{r_k\}$ be drawn
from the memory buffer  $\mathbb{D}$. Assume the variance of $\{r_k\}$ is upper bounded by a finite positive number $\mathbb{B}$, i.e., $var[r_k] \leq \mathbb{B}$. Further define a constant $\psi$ as,
\begin{equation}\label{Eq:psi}
    \psi = (\frac{\gamma - 1}{\gamma^N - 1})^2(\frac{\gamma^{2N} - 1}{\gamma^2 - 1}).
\end{equation}
Then the  upper bounds of the variances of the two $Q$ value sequences,  ${var[Q_{i+1}]}$ and $var[\mathbb{Q}_{i+1}]$, are respectively described below,
\begin{equation}\label{Eq:TH1_Q}
    var[Q_{i+1}(s_{k},a_{k})] \leq \sum_{t=1}^{i+1}(\gamma^{t-1})^2\mathbb{B}, 
\end{equation}
\begin{equation}\label{Eq:TH1_QQ}
    var[\mathbb{Q}_{i+1}(s_{k},a_{k})] \leq \psi\sum_{t=1}^{i+1}(\gamma^{t-1})^2 \mathbb{B}.
\end{equation}

{\textbf {Proof.}} We prove by mathematical induction. First let $i = 0$. Based on Equation (\ref{Eq:covrQeq0}), Equations (\ref{Eq:varQ}) and (\ref{Eq:varQQ}) become:

\begin{equation}
\begin{aligned}
    var[Q_{1}(s_{k},a_{k})] &= var[r_k], \\
    var[\mathbb{Q}_{1}(s_{k},a_{k})] &= var[r']. 
\end{aligned}
\end{equation}

First note that $var[Q_{1}(s_{k},a_{k})]\leq\mathbb{B}$ as $var[r_k]\leq\mathbb{B}$.

Then from Equation (\ref{Eq:r'}), $var[r']$ can be written as:
\begin{equation}
\begin{aligned}
        &var[r'] = var[\frac{\gamma - 1}{\gamma^N - 1}\sum_{t=k}^{k+N-1}\gamma^{t-k} r_t ] \\
        &= (\frac{\gamma - 1}{\gamma^N - 1})^2 var[\sum_{t=k}^{k+N-1}\gamma^{t-k} r_t] \\
        &= (\frac{\gamma - 1}{\gamma^N - 1})^2 \sum_{l=k}^{k+N-1}\sum_{t=k}^{k+N-1}(\gamma^{t-k})^2cov(r_t,r_l) \\
        &= (\frac{\gamma - 1}{\gamma^N - 1})^2 \sum_{t=k}^{k+N-1}(\gamma^{t-k})^2var[r_t] \\
        &\leq (\frac{\gamma - 1}{\gamma^N - 1})^2(\frac{\gamma^{2N} - 1}{\gamma^2 - 1}) \mathbb{B},
\end{aligned}
\end{equation}
where in the above, the equation above the inequality is obtained from applying Lemma 1. 


We thus have,
\begin{equation}
\begin{aligned}
    var[\mathbb{Q}_{1}(s_{k},a_{k})] \leq \psi\mathbb{B}. 
\end{aligned}
\end{equation}
This shows that the theorem holds for $i = 0$.


Then Assume that Equation (\ref{Eq:TH1_Q}) and (\ref{Eq:TH1_QQ}) hold for $i = l - 1$, where $l = 1, 2, ...$. Then, for $i = l$, we have
\begin{equation}
\begin{aligned}
    var[Q_{l + 1}(s_{k},a_{k})]  &= var[r_k] + var[\gamma Q_{l}(s_{k+1},a_{k+1})] \\
    &\leq \mathbb{B} + \gamma^2\sum_{t=1}^{l}(\gamma^{t-1})^2\mathbb{B} \\
    &= \sum_{t=1}^{l+1}(\gamma^{t-1})^2\mathbb{B} .\\
\end{aligned}
\end{equation}
Additionally, 
\begin{equation}
\begin{aligned}
    var[\mathbb{Q}_{l+1}(s_{k},a_{k})] &= var[r'] + var[\gamma \mathbb{Q}_i(s_{k+1},a_{k+1})] \\
    &\leq \psi\mathbb{B} + \gamma^2\psi\sum_{t=1}^{l}(\gamma^{t-1})^2\mathbb{B} \\
    &= \psi\sum_{t=1}^{l+1}(\gamma^{t-1})^2\mathbb{B}.  \\
\end{aligned}
\end{equation}

Thus Theorem 1 holds.

\begin{figure}[h]
\includegraphics[width=210pt]{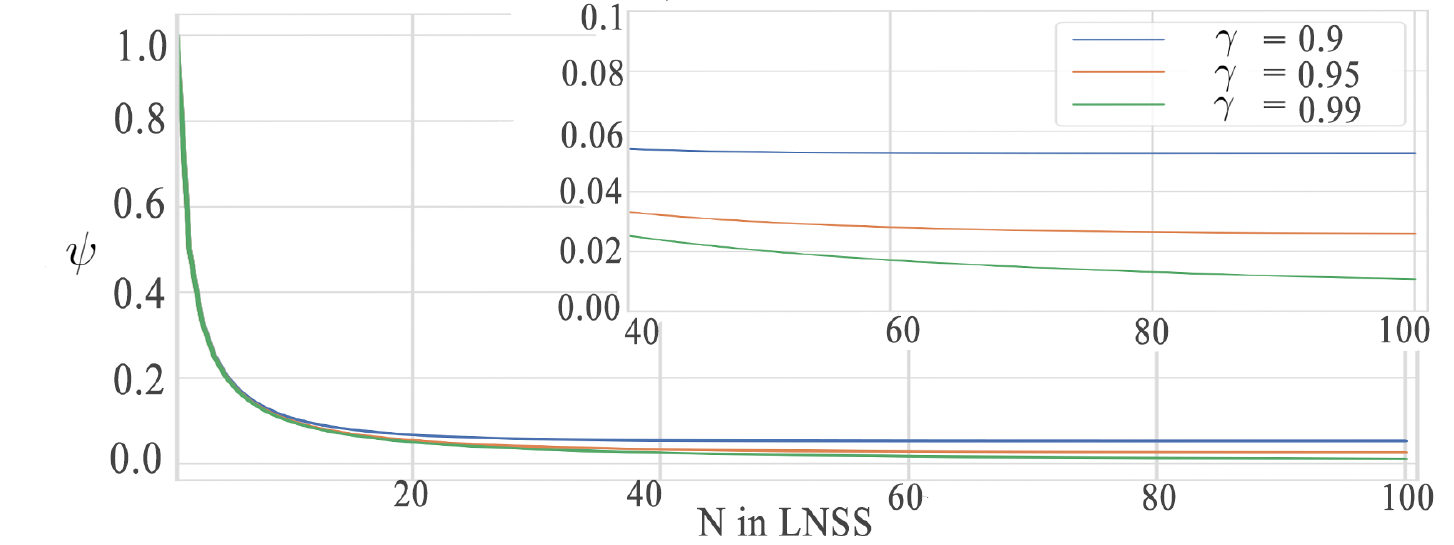}\
\caption{ Variance discount factor $\psi$ in Equation (\ref{Eq:psi}).}
\label{Fig.psi}
\end{figure} 
 
{\textbf {Remark 2}}. We now provide some insights based on the variance analysis above. 

1) Given $\psi$ in Equation (\ref{Eq:psi}), i.e., $\psi = (\frac{\gamma - 1}{\gamma^N - 1})^2(\frac{\gamma^{2N} - 1}{\gamma^2 - 1})$, it follows that for  large $N$, $\psi = {(\gamma - 1)}^2(\frac{ - 1}{\gamma^2 - 1})= \frac{1-\gamma}{1+\gamma}$. 

2) Furthermore, by the following identifies, $\gamma^2 - 1 = (\gamma - 1)(\gamma + 1)$ and $\gamma^{2N} - 1 = (\gamma^N - 1)(\gamma^N + 1)$, we have that $ \psi =
    (\frac{\gamma - 1}{\gamma + 1})(1 + \frac{2}{\gamma^N - 1})$.
Therefore, $\psi$ 
decreases exponentially (refer to Figure \ref{Fig.psi}).

3) From inspecting Equations (\ref{Eq:TH1_Q}) and (\ref{Eq:TH1_QQ})
we can see a clear advantage of using long $N$-steps in LNSS over the typical reward $r_k$.  




\section{Experiments and Results}
\begin{figure*}[ht]
    \centering
    \includegraphics[width=430pt]{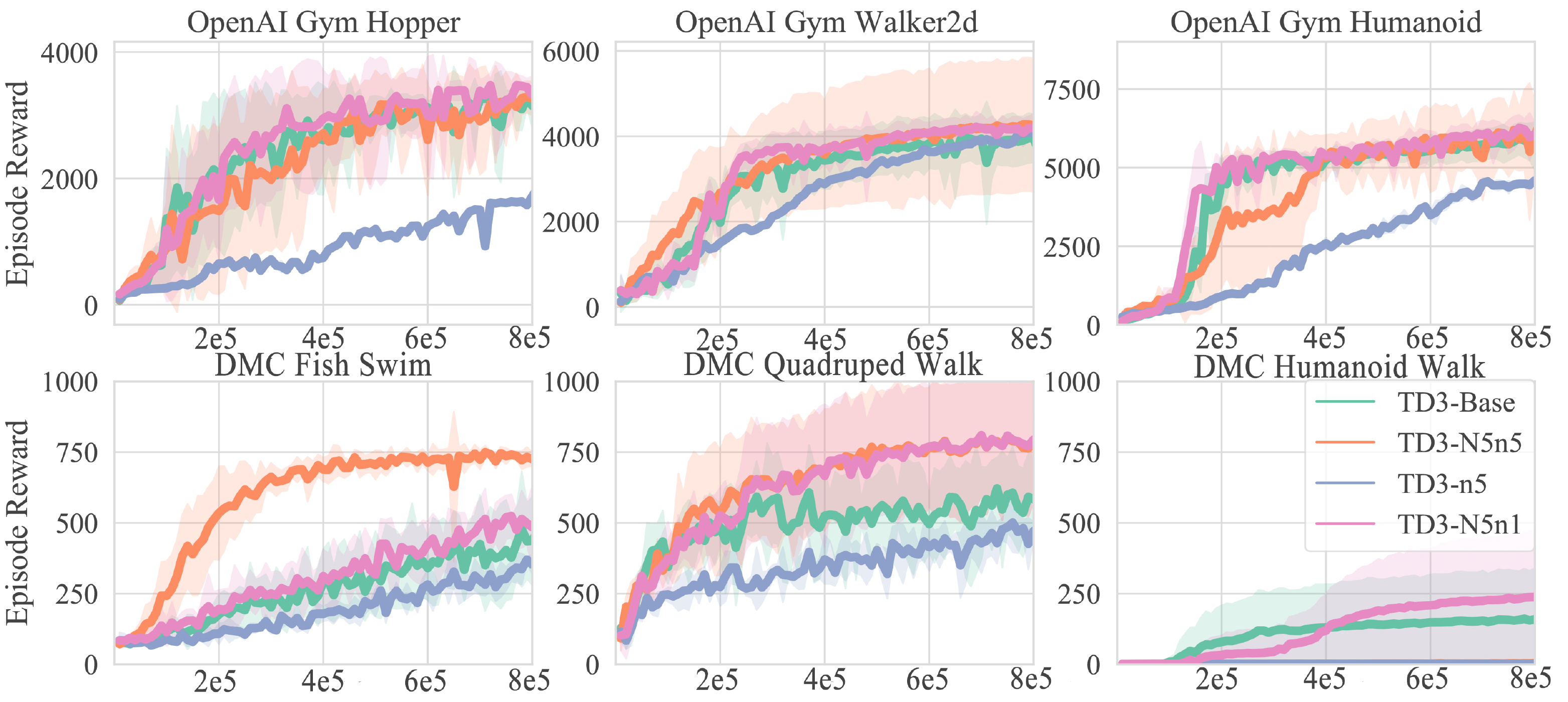}\
    \caption{Systematic evaluation of LNSS using several challenging  continuous control tasks in OpenAI Gym and DMC. The shaded regions represents half a standard deviation of the average  evaluation over 5 trials. The x-axis of the plots is the number of steps.}
    \label{fig:total Result}
\end{figure*}

We now provide a comprehensive evaluation of our proposed LNSS on top of TD3 by measuring its performance on several challenging benchmarks in  GYM and DMC.  To allow for reproducible comparison, we use the original tasks of hopper, walker2d, and humanoid in GYM; and humanoid walk, quadruppled walk, and fish swim in DMC with no modifications to the environments. 
Note that, in  LNSS TD3 learning, we elevated reward stored in the temporary buffer $\mathbb{D}'$ by 1.5 and 2, and lower bounded it by 0, for hopper and walker2d in GYM, respectively,  to keep the reward positive semi-definite.  In evaluation for comparisons, however, we follow the typical  procedure used previously \cite{duan2016benchmarking} 
by using the same original rewards in the original task environments to report all results in this study.

In reporting evaluation results below, we use the following short-form descriptions. 

1) ``TD3-Base'' denotes the  original TD3 and its implementation code from \cite{fujimoto2018addressing}. It is used in all evaluations of $n$-step methods and  LNSS.

2) ``TD3-n5"  denotes the $n$-step TD3 as in  Equation (\ref{Eq:n_step}) with $n=5$.

3) ``TD3-N5n5"  denotes the n-step TD3 as in the above 
but the n-step reward in  Equation (\ref{Eq:n_step})  is replaced by LNSS reward in Equation (\ref{Eq:r'})  with $N=5$.

4) ``TD3-N5n1"  denotes the original TD3 with LNSS reward $r'$  computed from Equation (\ref{Eq:r'}) using $N=5$, and with the Bellman Equation (\ref{Eq:Q}) computed with $n=1$. Similarly,  "TD3-N50n1" and "TD3-N100n1" use N=50 and N=100, respectively. 

5) ``Mean reward"  denotes the mean reward method \cite{yuan2019novel} as in Equation (\ref{Eq:mean_reward}) with $n = 100$.

 Details of the implementation, training, and evaluation processes are provided in Appendix \ref{appendix:addtional implementation}

\subsection{Baseline Studies}
We perform a baseline study by comparing LNSS on top of TD3 with the original TD3 (TD3-base)  \cite{fujimoto2018addressing},  $n$-step TD3  (TD3-n5)  and the mean reward method \cite{yuan2019novel}.

The results are presented in Figure \ref{fig:total Result} and Table \ref{table:ablation study}.  For all six tasks evaluated, with LNSS reward, TD3-N5n5 and TD3-N5n1 (orange line and pink line, respectively) significantly outperform TD3-n5 (blue line) and outperform or match TD3-base (green line)  both in terms of final reward and learning speed. In terms of elvaluation by coefficient of variation (CV), TD3-n5 has the lowest CV but its final reward is far behind those enabled by LNSS.  Overall, implementations using LNSS are behind   most of the top performances in terms of  high average reward, and low CV as shown in Table \ref{table:ablation study}.


\begin{table*}[ht]
\small
\begin{tabular}{|l|llllll|llllll|}
\hline
 & \multicolumn{6}{c|}{GYM}  & \multicolumn{6}{c|}{DMC}                             \\ \cline{2-13} 
 & \multicolumn{2}{c|}{Hopper}    & \multicolumn{2}{c|}{Humanoid}   & \multicolumn{2}{c|}{Walker2d}     & \multicolumn{2}{c|}{Fish Swim}     & \multicolumn{2}{c|}{Humanoid Walk}   & \multicolumn{2}{c|}{Quadruped Walk} \\ \cline{2-13}
& Reward & CV & Reward & CV & Reward & CV & Reward & CV & Reward & CV & Reward & CV \\ \hline
TD3-Base  & 3185.52 &0.085  & 5884.46 &0.092 & 3922.09 &0.14 & 431.23 &0.34 & 157.75 &1.04 & 574.25 &0.25          \\

TD3-n5   & 1638.44 &\textbf{0.031}  &  4497.75 &\textbf{0.025}  &  4031.18 &\textbf{0.026}  &  346.32 &\textbf{0.11}  & 1.94 &\textbf{0.077}   &  469.90 &\textbf{0.14}    \\ 

TD3-N5n5   & 3244.21 &0.094  &   5974.76 & \textbf{0.022}  &  4254.49 &0.34  & \textbf{731.67} &\textbf{0.033}  &  4.00 &0.44   &  779.88 &0.27  \\

TD3-N5n1   & \textbf{3403.97} &\textbf{0.052} & \textbf{6061.21} &0.080 & 4164.08 &\textbf{0.066}  & 499.67 &0.18   & \textbf{235.14} &0.89 & \textbf{785.16} &0.27 \\

TD3-N50n1  & \textbf{3443.71} &0.057   &  \textbf{6223.58} &0.046  &  \textbf{4542.72} &0.072   &  \textbf{576.65} &\textbf{0.12}  & \textbf{266.58} &\textbf{0.19}   &   \textbf{821.33} &\textbf{0.10}    \\

TD3-N100n1  & \textbf{3373.94} &\textbf{0.052}  &  \textbf{6061.71} &\textbf{0.014} &  \textbf{4541.77}&\textbf{0.045}  &  \textbf{618.58} &\textbf{0.12} &  \textbf{285.78} &\textbf{0.24}  &   \textbf{783.18} &\textbf{0.21}    \\

Mean reward  & 2103.48 &0.515  &  5321.36 &0.054  &  \textbf{4554.81} &0.22 &  214.33 &0.28  &  65.61 &1.43  &   554.50 &0.22    \\
\hline
\end{tabular}
\caption{Summary of performances of the compared algorithms based on the  last 5 evaluations for five different trials.  Top 3  performances  are boldfaced for average reward and  coefficient of variation (CV), respectively.}
\label{table:ablation study}
\end{table*}

\begin{figure}[h]
    \centering
    \includegraphics[width=200pt]{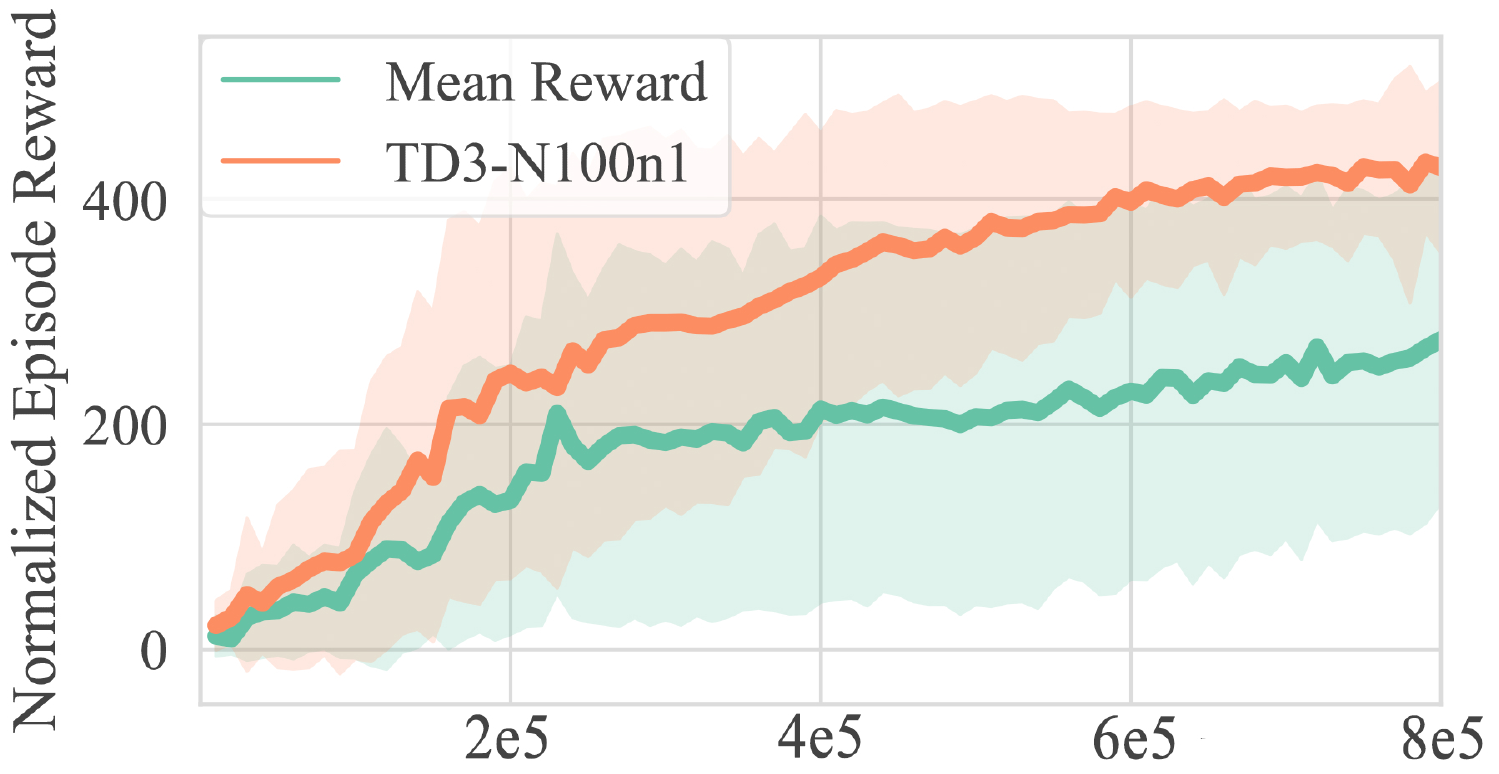}\
    \caption{Performance comparison  between LNSS (N=100) and mean reward method (n=100). All 6 tasks are considered for the two algorithms. Specifically, for each algorithm and each task, episode rewards are normalized to $[0,500]$. Then episode rewards for all six tasks for each algorithm are plotted together in one color. 
    The shaded regions represent half a standard deviation of the average evaluation over 5 episodes. The x-axis of the plots is the number of steps. Detailed results for individual tasks are shown in Appendix \ref{appendix:vs mean reward}.}
    \label{fig:normalization}
\end{figure}

Then, in Figure \ref{fig:normalization}, we compare performance of  TD3-N100n1 with  mean reward method using long reward sequences ($n = 100$ in Equation (\ref{Eq:mean_reward})). For better depiction of results of all tasks in  both GYM and DMC, we normalize each task reward into a scale of $[0,500]$ to show overall performance in perspective. The detailed results of individual tasks are provided in Appendix \ref{appendix:vs mean reward} Figure \ref{fig:normalization Full}. Overall, with LNSS reward, TD3-N100n1 outperforms the mean reward method in all tasks in terms all measures (final reward, CV, and learning speed).



\begin{figure}[htbp]
    \centering
    \includegraphics[width=200pt]{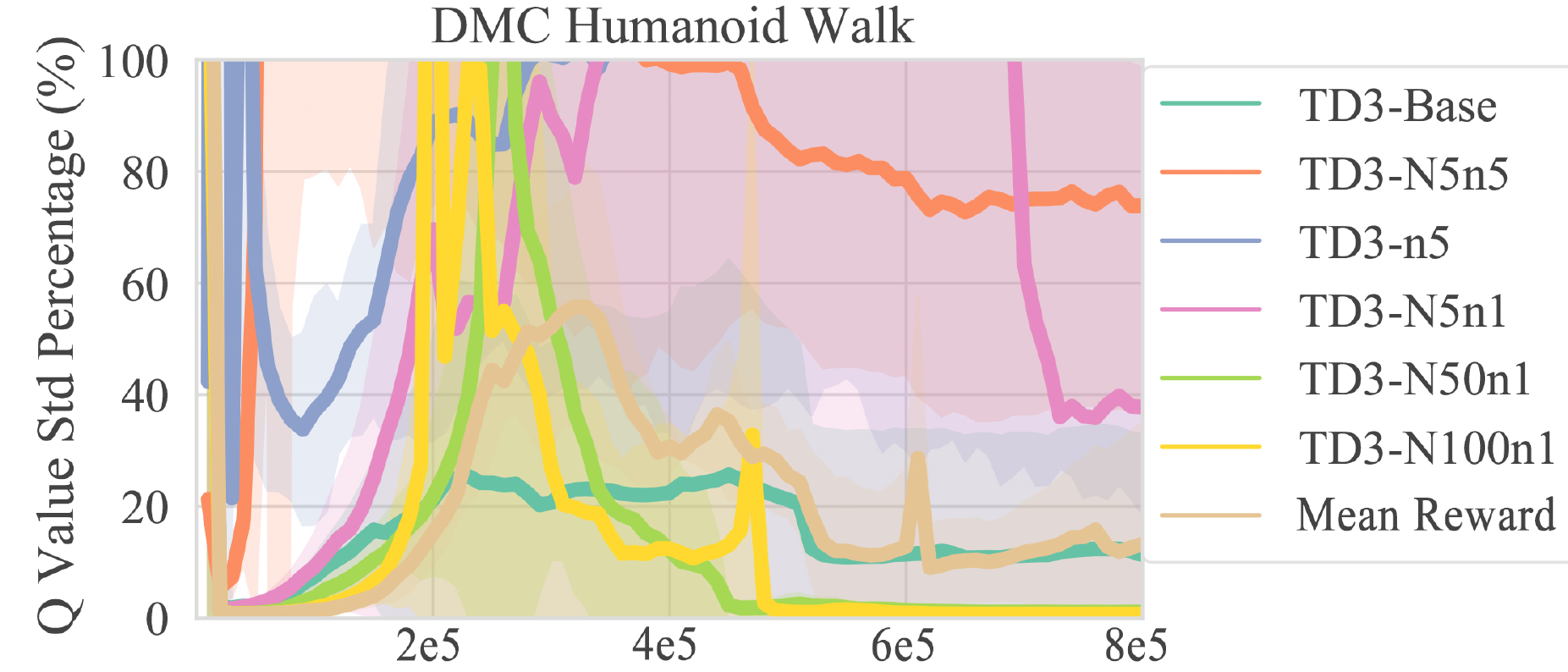}\
    \caption{Q value Std percentage of all tested algorithms in DMC huamnoid walk task. The x-axis  is the number of steps. For detailed results of other DMC tasks, refer to Figures \ref{fig:variance full} in Appendix \ref{appendix:variance}.}
    \label{fig:variance percentage}
\end{figure}

Furthermore in Figure \ref{fig:variance percentage}, we show the $Q$ value variance percentage performance of the algorithms based on the measure of $\frac{std(Q)}{Q}\%$. Here we show the humanoid-walk task result since it is considered the most challenging in our evaluations. The detailed results for individual tasks in DMC  
are shown in Appendix \ref{appendix:variance} Figure \ref{fig:variance full}. 
Due to early termination inherent in GYM, it interferes with LNSS (especially for large $N$)  to appropriately utilize the entire trajectory of desired  $N$ steps. We thus report results on  investigations of length $N$ in LNSS using DMC tasks. 
Figure \ref{fig:variance percentage} provides an empirical validation for our theoretical analysis on bounds of variances  of the  $Q$  value, namely,  with large $N$ (yellow and green) or at late stage of learning, $Q$ value variances become smaller as the upper bound of $Q$ value variance reduces significantly according to  Equation (\ref{Eq:psi}) in Theorem 1.

\begin{figure}[h]
\centering
\includegraphics[width=200pt]{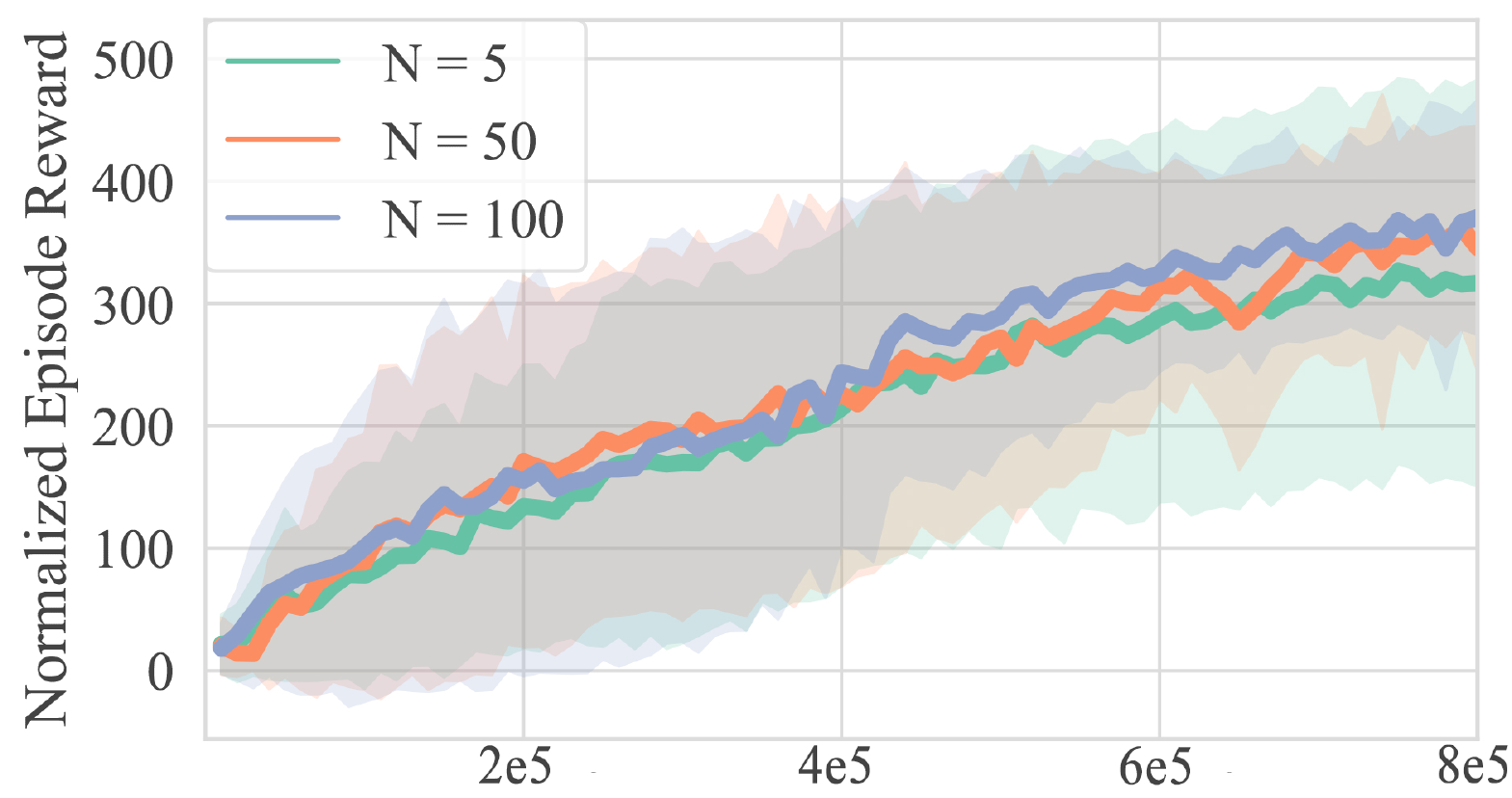}\
\caption{Episode rewards for the 3 tasks in DMC by LNSS with $n=1$ but different $N$ ($N = 5 , 50, 100$) in  Equation (\ref{Eq:r'}). The Episode rewards   for each task are normalized to [0, 500]. The shaded regions represent half a standard deviation of the average evaluation scores over 5 trials. The x-axis  is the number of steps. Additional details are provided in Appendix \ref{appendix:N step}.}

\label{fig:Nstep}
\end{figure}

\subsection{Evaluation of LNSS}

In Figure \ref{fig:Nstep}, we show the effect of different choices of $N$ $(N=5,50,100)$ in our proposed LNSS. Comparisons are performed  using the same set of  hyperparameters except that $N$ is different in all experiments. Figure \ref{fig:Nstep} depicts performance curves for DMC tasks, we normalize each task reward into a scale of $[0,500]$ to show overall performance in perspective. The full detailed results for  the 6 individual tasks  are provided in Appendix \ref{appendix:N step}, Figure \ref{fig:Nstep Full}. The LNSS with $N=50$ and $N=100$ outperform LNSS with $N=5$ in terms of final reward, learning speed, and CV. The largest $N=100$ in LNSS match the final reward and CV for  $N=50$, but $N=1000$ has resulted in fastest learning speed among all the compared. 
Additionally, inspecting Figure \ref{fig:Nstep Full} in Appendix \ref{appendix:N step} we easily realize the trend that without an increase in CV, the more complex a task is,  such as fish-swim and humanoid-walk, the more effective a large $N$ is as measured by learning speed and final reward. 
For relatively simple tasks such as quadruped-walk, $N = 50$ and $N = 100$ show comparable performances.

\subsection{Ablation Studies}
We perform ablation studies to understand the contribution of LNSS. Respective results are summarized in Table \ref{table:ablation study} and Figure \ref{fig:total Result} where we compare performances by removing each component from LNSS. 

We first show how directly applying $n$-step method  (TD3-n5) to a DRL algorithm may hurt its performance. By inspecting  Figure \ref{fig:total Result} and Table \ref{table:ablation study}, performances of TD3-n5 (blue line) and TD3-Base (green line), we see that the $n$-step method  does help reduce CV but hurts learning speed and final reward.

We then show how our LNSS can improve TD3-Base. By inspecting  Figure \ref{fig:total Result} and Table \ref{table:ablation study}, performances of TD3-N5n1 (pink line) and TD3-Base (green line), LNSS has improved the original TD3 performance in terms of  all measures.

Then we show how our LNSS can help $n$-step methods for solving complex problems in GYM and DMC. By inspecting performances of TD3-N5n5 (orange line) and TD3-n5 (blue line) in Figure \ref{fig:total Result} and Table \ref{table:ablation study}, LNSS has significantly improved TD3-n5 performance in final reward and learning speed with a little compromised CV.

Finally  we compare performance of LNSS using the same $N$ in LNSS but different $n$ in $Q$ value update, specifically we inspect results from  TD3-N5n1 and TD3-N5n5. 
From  Figure \ref{fig:total Result} and Table \ref{table:ablation study}, the two implementations result in similar performance most of the time. But $n=5$ in $Q$ value update can benefit some environment such as Fish Swim. However, it doesn't help for more complex problems such as humanoid-walk. 
Overall, TD3-N5n1 has more stable performance in each task.

\subsection{Limitation of This Study.} 
Here we introduce a new multi-step method, the LNSS, to effectively utilize longer future steps than those used in existing methods to estimate the $Q$ value with an aim of reducing variances in learning.  By imposing LNSS on top of TD3, we demonstrate improved performance in terms of final reward, learning speed, and variance in $Q$ value. These benefits are especially enlarged in those very complex tasks. However, there exist two limitations. 1) The LNSS requires positive semi-definite reward values in a long trajectory. Large negative reward may wash out the significance of good positive rewards. However, this limitation can be easily fixed by elevated reward stored in the temporary buffer $\mathbb{D}'$ by a positive constant, and lower bounded it by 0. 2) Early termination is another limitation, which mostly affects  LNSS in GYM environments (hopper, humanoid, walker2d). Different from DMC tasks, GYM allows an environment to terminate before reaching the maximum time steps. As such, in early training stage, only 10 or 20 steps can be used to compute LNSS which diminish the power of  large $N$ (such as $50$ or $100$). To resolve this limitation,  adjusting task settings in environments is required.

\section{Discussion and Conclusion}
1) In this work, we introduce a novel $N$-step method, the LNSS, that is easy to implement with low computational cost, and applicable to any value-based or policy gradient DRL algorithms. It has been shown nearly consistently outperforming similar algorithms in solving complex benchmark tasks in terms of  performance score,  coefficient of variation, and convergence speed. 2) We provide a theoretical analysis to show that  LNSS reduces the upper bound  of variance in $Q$ value 
exponentially  from respective single step methods. 
3) We empirically demonstrate 
the performance of   LNSS on top of TD3
in complex GYM and DMC 
benchmark environments that have been challenging for existing methods to obtain good results. 
Our results suggest that LNSS is a  promising tool for improving learning speed,  learning performance score, and reducing learning variance. Further investigation on how to maximize the benefit of selecting an optimal reward length $N$, and how to take advantage of a different $n$ in $Q$ value update are exciting questions to be addressed in future work. Additionally, we look forward to applying LNSS to other DRL algorithms in different tasks.  

\clearpage

\begin{appendices}
\section{Additional Implementation Details}
\label{appendix:addtional implementation}
We use PyTorch for all implementations. All results were obtained  using our internal server consisting of AMD Ryzen Threadripper 3970X Processor, a desktop with Intel Core i7-9700K processor and a desktop with AMD Ryzen 7 3800XT processor. 
Our implementation code is detailed and provided in Appendix. {\ref{appendix:code}}. 

\textbf{Training Procedure}. 

A episode is initialized by resetting the environment, and terminated at max step $T=1000$ or early termination if criteria are met depending on specific tasks. A trial is a complete training process contains a series of consecutive episodes. Each trial is run for a maximum $8\times10^5$ time steps with evaluations at every $1\times10^4$ time steps. Each task is reported over 5 trials where the environment and the network were initialized by 5 mother random seeds, $(0-4)$ in this study.

For each training trial, to remove the dependency on the initial parameters of a policy, we use a purely exploratory policy for the first 8000 time steps (start timesteps). Afterwards, we use an off-policy exploration strategy, adding Gaussian noise $\mathcal{N}(0,0.1)$ to each action. 

\textbf{Evaluation Procedure}.

Every $1\times10^4$ time steps training, we have a evaluation section and each evaluation reports the average reward over 5 evaluation episodes, with no exploration noise and with fixed policy weights. The random seeds for evaluation are different from those in training which are shown in the following section on Random Seed.

\textbf{Random Seed}. 

Each of the  tasks we used in this work was trained for 5 trials with 5 different random mother seeds $s_m$ (0,1,...,4) across all algorithms. Within each trial, evaluations were performed using seeds  $(s_m + 100)$. For distributed training, we implement 8 parallel actors to generate experiences and each actor $a_i, i = 0,1,2,...7$ has the seed $s_{a_i}$ as $s_{a_i} = s_m + i$. seed $s_{a_i}$ are  used in environment, Numpy, PyTorch for seed initialization.

\textbf{Network Structure and optimizer}.

The actor-critic networks in TD3 are implemented by feedforward neural networks with three layers of weights. Each layer has 256 hidden nodes with rectified linear units (ReLU)  for both the actor and critic. 
The input layer of actor has the same dimension as observation state. The output layer of the actor has the same dimension as action requirement with a tanh unit. Critic receives both state and action as input to THE first layer and the output layer of critic has 1 linear unit to produce $Q$ value. 
Network parameters are updated using Adam optimizer with a  learning rate of $10^{-3}$. After each time step $k$, the networks are trained with a mini-batch of a 256 transitions $(s,a,r,s')$, $(s,a,r',s')$ in case of LNSS, sampled uniformly from a replay buffer $\mathbb{D}$ containing the entire history of the agent.

\textbf{Policy update}. 

Target policy smoothing is implemented by adding $\epsilon \sim \mathcal{N}(0,0.2)$ to the actions chosen by the target actor network, clipped to $(-0.5, 0.5)$, delayed policy updates consists of only updating the actor and target critic network every $d$ iterations, with $d = 2$. While a larger $d$ would result in a larger benefit with respect to accumulating errors, for fair comparison, the critics are only trained once per time step, and training the actor for too few iterations would cripple learning. Both target networks are updated with \textbf{$\tau = 0.005$}.

The  TD3 used in this study is based on the paper \cite{fujimoto2018addressing} and the code from the authors (https://github.com/sfujim/TD3). The distributed learning process is based on \cite{clemente2017efficient}. 

\begin{table}[h]
\begin{tabular}{l|llll}
Hyperparameter           & Value                   &  &  &  \\ \cline{1-2}
Start timesteps          & 8000 steps              &  &  &  \\
Evaluation frequency     & 10000 steps             &  &  &  \\
Max timesteps            & 8e5 steps               &  &  &  \\
Exploration noise        & $\mathcal{N}(0,0.1)$    &  &  &  \\
Policy noise             & $\mathcal{N}(0,0.2)$    &  &  &  \\
Noise clip               & $\pm 0.5$               &  &  &  \\
Policy update frequency  & 2                       &  &  &  \\
Batch size               & 256                     &  &  &  \\
Buffer size              & 1e6                     &  &  &  \\
$\gamma$                 & 0.99                    &  &  &  \\
$\tau$                   & 0.005                   &  &  &  \\
Number of parallel actor & 8                       &  &  &  \\
LNSS-N                   & choose as results shows &  &  &  \\
LNSS-n                   & choose as results shows &  &  &  \\
Adam Learning rate       & 1e-3                    &  &  &  \\

\end{tabular}
\caption{TD3 + LNSS hyper parameters used for the GYM and DMC benckmark tasks}
\label{table:hyperparam}
\end{table}

{\textbf {Distributed Learning Procedure}}. Distributed Learning is widely used in DRL algorithms \cite{schulman2017proximal,barth2018distributed,hessel2018rainbow} for speeding up experience gathering. Note from Equation (\ref{Eq.Loss}) and (\ref{Eq.Actor_up}) that updating the actor and the critic relies on sampling from some state distributions $p_{\pi}(s)$. We can parallelize this process by using a distributed process of multiple independent actors, each writing to the same memory buffer. Samples of experiences  from the memory buffer can then be used in learning.

To speed up the learning process, we use a distributed implementation to parallelize computation. In the style of \cite{clemente2017efficient}, we use a centralized agent with several workers operating in parallel. Each worker loads the most recent policy, interacts with the environment, and sends its observations to the central agent. Given the computing resource, $8$ workers were implemented in this study. All algorithm hyper-parameters are summarized in Table \ref{table:hyperparam}.

\section{Code Details}
\label{appendix:code}

Here we provide our code details in both algorithm box and real code implementation centered on LNSS. The TD3 implementation was as described in the above.

\begin{algorithm}
\caption{Long $N$-step Surrogate Stage (LNSS) Reward }
\label{alg:algorithm}
\textbf{Given}:
\begin{itemize}
  \item an on/off-policy RL algorithm $\mathbb{A}$,          e.g PPO,TD3,DQN
  \item $N$-step number  $N$
  \item $n$-step update  $n$
  \item an experience buffer $\mathbb{D}$
  \item a temporary experience buffer $\mathbb{D}'$ with size $N$
  \item Total training episode  $\mathbb{T}$
\end{itemize}
\textbf{Initialize}: $\mathbb{A}$, $\mathbb{D}$, $\mathbb{D}'$ 
\begin{algorithmic}[1] 
\FOR{episode = 1, $\mathbb{T}$}
\STATE Reset initialize state $s_0$, $\mathbb{D}'$
\FOR{k = 0, $T$}
    \STATE Choose an action $a_k$ based on current state $s_k$ and learned policy from $\mathbb{A}$.
    \STATE Execute the action $a_k$ and observe a new state $s_{k+1}$ with reward signal $r_k$
    \STATE Store the transition $(s_k,a_k,r_k,s_{k+1})$ in $\mathbb{D}'$
    \IF {$k + N - 1 \leq T$}
        \STATE Get earliest memory $(s_0',a_0',r_0',s_1')$ in the $\mathbb{D}'$
        \STATE Calculate $r'$ based on Equation (\ref{Eq:calc_r'1})
        \STATE Store the transition $(s_0',a_0',r',s_1')$ in $\mathbb{D}$
        \STATE Clear transition $(s_0',a_0',r_0',s_1')$ in the $\mathbb{D}'$
    \ELSE
        \STATE Repeat step 8 to 11 and Calculate $r'$ based on Equation (\ref{Eq:calc_r'2})
    \ENDIF
    \STATE using $r'$ to perform $n$-step of optimization using $\mathbb{A}$ and mini-batch data from $\mathbb{D}$
\ENDFOR
\ENDFOR

\end{algorithmic}
\end{algorithm}

LNSS implementation code used in obtaining all the results in this study is provided below.

\begin{lstlisting}

#import deque for tempory buffer D'
from collections import deque
#Initialize replay buffer with:
#state and action dimension & buffer size
replay_buffer = ReplayBuffer(state_dim, action_dim, int(buffer_size))

#Initialize tempory buffer D'
exp_buffer = deque()

#LNSS N factor as Equation (12)
N_step_number = args.N_step

#Initialize environment
state, done = env.reset(), False
#Initialize episode timesteps 
episode_timesteps = 0
#Start training steps until max timesteps
for t in range(int(args.max_timesteps)):
	# Select action according to policy from RL algorithms (TD3)
	action = policy.select_action(np.array(state))
	# Perform action
	next_state, reward, done, _ = env.step(action) 
	# Store transition in temporary buffer D'
	exp_buffer.append(state, action, reward,next_state)
	
	
	#LNSS computation 
	if len(exp_buffer) >= N_step_number:
	    #get n step state data for n-step update in Equation (13)
		_, _, _, next_state_1,done_1 = exp_buffer[int(args.n_update-1)]
		#get earliest data tuple in D' and clear them in D'
		state_0, action_0, reward_0,_,_ = exp_buffer.popleft()
		#start to calculate r' 
		discounted_reward = reward_0
		gamma = args.discount
		for (_, _, r_i, _, _) in exp_buffer:
			discounted_reward += r_i * gamma
			gamma *= args.discount
			
		#apply discounted factor to reward
		ds_factor = (args.discount - 1)/(gamma - 1)
		#final r' as
		discounted_reward = ds_factor * discounted_reward
		#store data in memory buffer D
		replay_buffer.add(state_0, action_0, discounted_reward,next_state_1)

    #next time step:
    
    episode_timesteps +=1
    state = next_state
    
    #RL algorithm training based on D and batch_size:
    policy.train(replay_buffer, args.batch_size)
	#if reaches the end of episode or early termination	
	if done:
		
		#compute r' based on rest of experiences remaining in buffer 
		while len(exp_buffer) != 0:
			len_buffer = len(exp_buffer)
			
			if len_buffer >= int(args.n_update):
				_, _, _, next_state_1,done_1 = exp_buffer[int(args.n_update-1)]
			else:
				_, _, _, next_state_1,done_1 = exp_buffer[len_buffer - 1]
				
			state_0, action_0, reward_0,next_state_0,done_bool_0 = exp_buffer.popleft()
			discounted_reward = reward_0
			gamma = args.discount
			for (_, _, r_i, _, _) in exp_buffer:
				discounted_reward += r_i * gamma
				gamma *= args.discount
				
			#apply discounted factor to reward
			ds_factor = (args.discount - 1)/(gamma - 1)
			discounted_reward = ds_factor * discounted_reward	
			#store data in memory buffer D 
			replay_buffer.add(state_0, action_0, discounted_reward,next_state_1)	
		
		#clear the D' for make sure previous data not effect next episode
		exp_buffer.clear()
\end{lstlisting}

\section{Background}
\label{appendix:background}

Here we present some DRL algorithm details regarding signle step updates and $n$-step updates:

1) DQN \cite{mnih2013playing} is a well-known value based methods. It uses deep neural networks to approximate action value function  $Q_{\theta}(s_k,a_k)$ (short hand notation $Q(s_k,a_k)$), as in $Q$-learning, with parameter $\theta$. The parameterized action value function $Q(s_k,a_k)$ is updated iteratively to solve the Bellman equation:
\begin{equation}
    Q(s_k,a_k) = \mathbb{E}_{s\sim p_{\pi}, a \sim \pi}[r_k + \gamma \max_{a}Q(s_{k+1}, a)]
\end{equation}
where actions follow policy $\pi = arg \max_{a} Q(s_k,a)$ and the state distribution $p_{\pi}(s)$ depends on the parameters in policy.

Let  $y = \mathbb{E}_{s\sim p_{\pi}, a \sim \pi}[r_k + \gamma \max_{a}Q(s_{k+1}, a)]$ be the target function to be used in minimizing the Bellman loss:
\begin{equation}\label{Eq.DQN_Loss}
     L\left(\theta\right) =\mathbb{E}_{s\sim p_{\pi}, a \sim \pi} [(y - Q(s_{k}, a_{k}))^2]
\end{equation}
Weight updates are according to gradient descent to minimize the loss function $L(\theta)$ by differentiating loss $L(\theta)$ with respect to the weights:
\begin{equation}
\begin{aligned}
    \nabla_{\theta} L\left(\theta\right) &=\mathbb{E}_{s\sim p_{\pi}, a \sim \pi_i} [(r_k \\
    &+ \gamma \max_{a}Q(s_{k+1}, a) \\
    &- Q(s_k, a_k))\nabla_{\theta} Q(s_k, a_k)].
\end{aligned}
\end{equation}

2) Rainbow \cite{hessel2018rainbow} integrate the $n$-step method as Equation. (\ref{Eq:n_step}), they have the $n$-step reward as $R_n = \sum_{t=k}^{t=k+n-1} \gamma^{t-k} r_{t}$ and use this $n$-step reward into the target function as:
\begin{equation}
    y = \mathbb{E}_{s\sim p_{\pi}, a \sim \pi}[R_n + \gamma^n \max_{a}Q(s_{k+n}, a)
\end{equation}
and this target will send to loss function in Equation (\ref{Eq.DQN_Loss}).

\begin{figure*}[h]
    \centering
    \includegraphics[width=500pt]{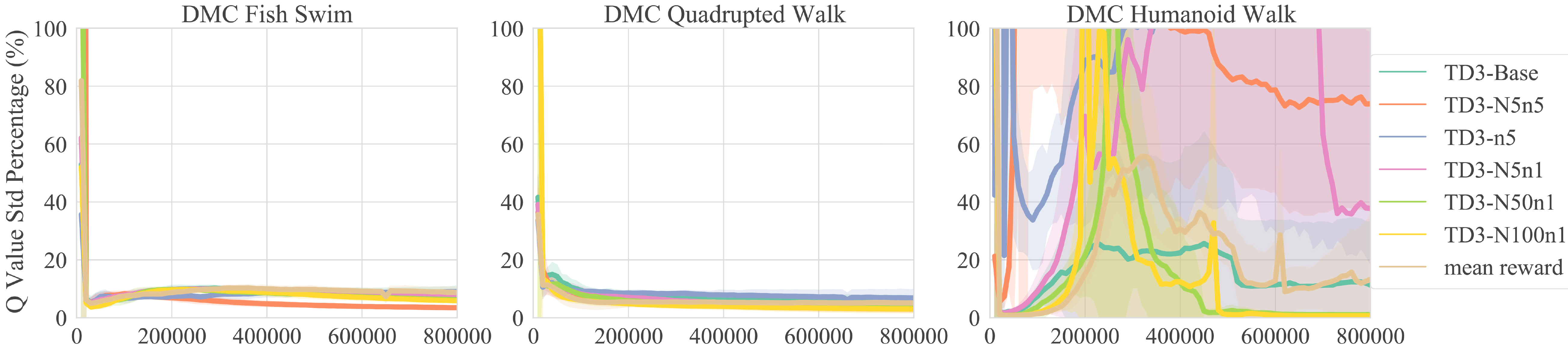}\
    \caption{Variance percentage of different algorithms over all tasks. The shaded region represents half a standard deviation of the average evaluation over 5 trials.}
    \label{fig:variance full}
\end{figure*}

3) PPO \cite{schulman2017proximal} is an on-policy policy gradient method that learns the state value function $V(s_k)$. Based on  a fixed length ($T$) trajectory, in stead of directly use $V(s_k)$ in Bellman equation, 
it uses a truncated version of a generalized advantage estimation (GAE) of $A_k$ \cite{schulman2015high}:
\begin{equation}
\begin{aligned}
\hat{A}_{k} &=\delta_{k}+(\gamma \lambda) \delta_{k+1}+\cdots+\cdots+(\gamma \lambda)^{T-k+1} \delta_{T-1}, \\
\delta_{k} &=r_{k}+\gamma V\left(s_{k+1}\right)-V\left(s_{k}\right),
\end{aligned}
\end{equation}
where $\lambda$ is the GAE parameter \cite{schulman2017proximal} to compromise between estimation variance and bias.

4) D4PG \cite{barth2018distributed} is based on DDPG and they utilizes $n$-step returns when estimating the TD error and replacing the Bellman operator with $n$-step variant:
\begin{equation}
\begin{aligned}
\left(\mathcal{T}_{\pi}^{N} Q\right)\left(\mathbf{x}_{0}, \mathbf{a}_{0}\right)&= r\left(\mathbf{x}_{0}, \mathbf{a}_{0}\right)+\mathbb{E}[\sum_{n=1}^{N-1} \gamma^{n} r\left(\mathbf{x}_{n}, \mathbf{a}_{n}\right) \\
&+\gamma^{N} Q\left(\mathbf{x}_{N}, \pi\left(\mathbf{x}_{N}\right)\right) \mid \mathbf{x}_{0}, \mathbf{a}_{0}]
\end{aligned}
\end{equation}
where $\left(\mathcal{T}_{\pi}^{N} Q\right)\left(\mathbf{x}_{0}, \mathbf{a}_{0}\right)$ is the distributional $Q$-value function they proposed in \cite{barth2018distributed} and the expectation is with respect to the $n$-step transition dynamics.

\section{Additional Detailed Results}

Here we provide additional detailed results to supplement those that we have reported in the paper.
This is a complete set of results for all experiments.

\begin{figure*}[h]
    \centering
    \includegraphics[width=450pt]{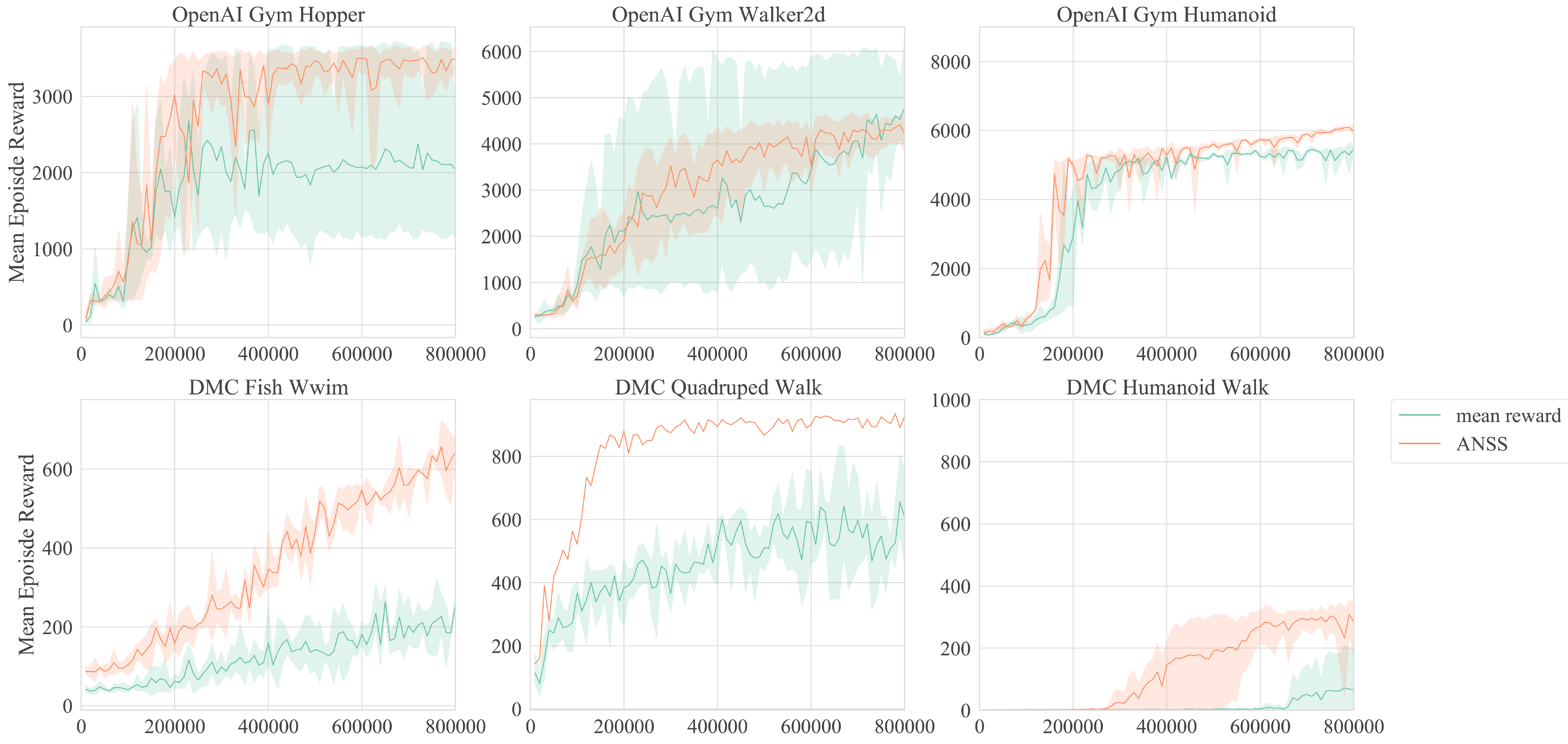}\
    \caption{Overall performance curves of LNSS and mean reward method over all tasks. The shaded region represents half a standard deviation of the average evaluation over 5 trials.}
    \label{fig:normalization Full}
\end{figure*}

\subsection{Variance analysis}
\label{appendix:variance}
In Figure \ref{fig:variance full}, we show the $Q$ value variance percentage performance of the compared algorithms based on the measure of $\frac{std(Q)}{Q}\%$ in DMC tasks. Across all algorithms, in Fish-Swim task, TD3-N5n5 (Pink line) has the lowest variance and for quadruped-walk and humanoid-walk TD3-N100n1 (yellow line) has the lowest variance. This shows that our LNSS methods can help algorithms to reduce variance in both single-step and $n$-step method. 

To investigate the effect of different choices of $N$ $(N=5,50,100)$ in our proposed LNSS, TD3-N100n1 (yellow line) has the lowest variance. Results of all 3 tasks provide an empirical validation for our theoretical analysis on bounds of variances  of the  $Q$  value, namely,  with large $N$ (yellow and light green) or at late stage of learning, $Q$ value variances become smaller as the upper bound of $Q$ value variance reduces significantly according to  Equation (\ref{Eq:psi}) in Theorem  1.

\subsection{LNSS vs. mean reward method}
\label{appendix:vs mean reward}
In Figure \ref{fig:normalization Full}, we compare performance of  TD3-N100n1 with  mean reward method using long reward sequences ($n = 100$ in Equation (\ref{Eq:mean_reward})) in all 6 tasks. The figure shows, with LNSS reward, TD3-N100n1 outperforms the mean reward method in all tasks in terms all measures (final reward, CV, and learning speed).


\subsection{Hyper-parameter Comparison}
\label{appendix:N step}

\begin{figure*}[ht]
    \centering
    \includegraphics[width=450pt]{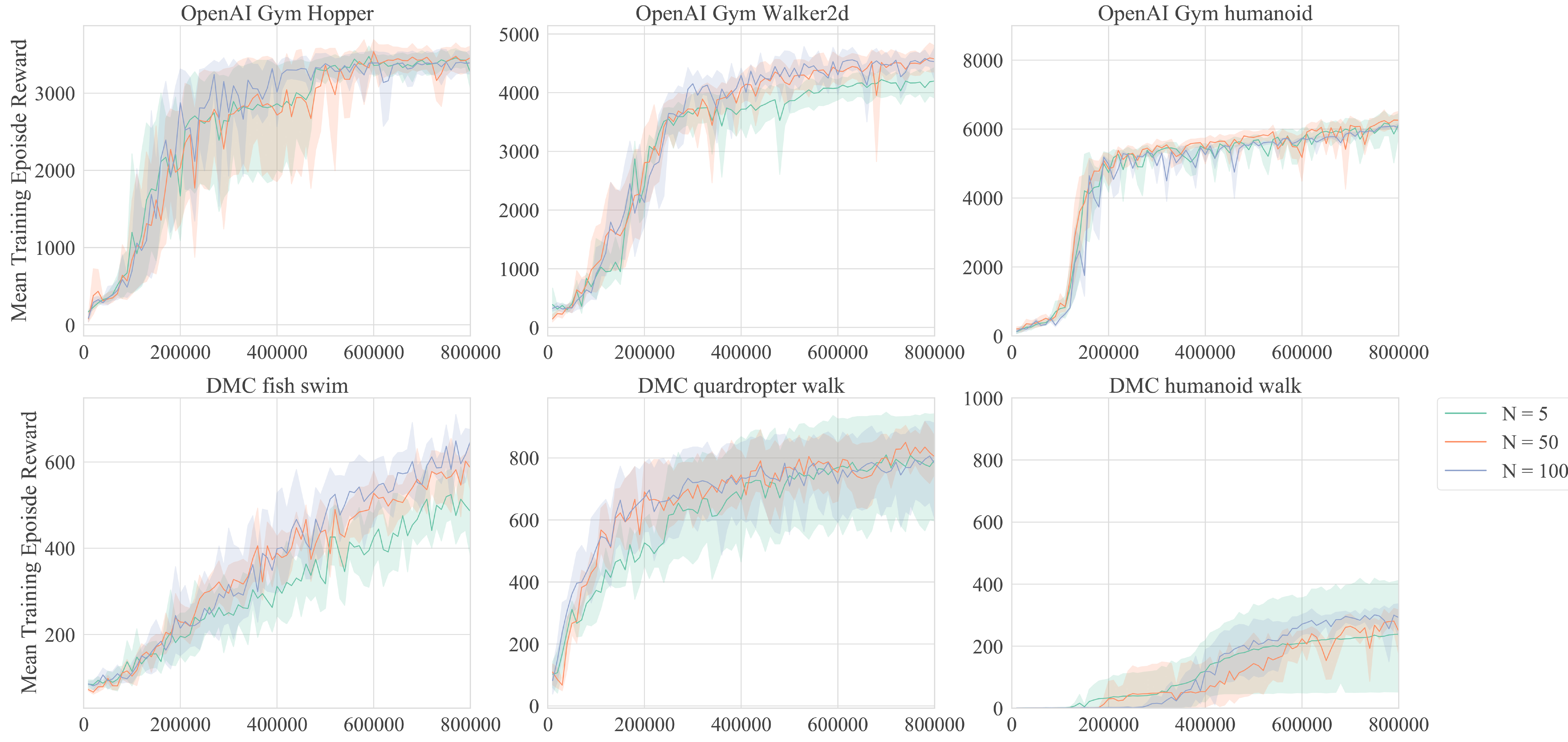}\
    \caption{Overall Performance curves of different $N$ factor over all tasks The shaded region represents half a standard deviation of the average evaluation over 5 trials.}
    \label{fig:Nstep Full}
\end{figure*}

In Figure \ref{fig:Nstep Full}, we show the effect of different choices of $N$ $(N=5,50,100)$ in our proposed LNSS. Comparisons are performed  using the same set of  hyperparameters except that $N$ is different in all experiments. The significance of large $N$ value varies task to task. In the DMC environment, $N=100$ outperforms others in reward and learning speed and matches others in CV. However, the improvement of $N = 100$ is limited in GYM environment. Due to early termination inherent in GYM, it interferes with LNSS (especially for large $N$)  to appropriately utilize the entire trajectory of desired  $N$ steps. 


\end{appendices}


\begin{thebibliography}{27}
\providecommand{\natexlab}[1]{#1}

\bibitem[{Anschel, Baram, and Shimkin(2017)}]{anschel2017averaged}
Anschel, O.; Baram, N.; and Shimkin, N. 2017.
\newblock Averaged-dqn: Variance reduction and stabilization for deep
  reinforcement learning.
\newblock In \emph{International conference on machine learning}, 176--185.
  PMLR.

\bibitem[{Barth-Maron et~al.(2018)Barth-Maron, Hoffman, Budden, Dabney, Horgan,
  Tb, Muldal, Heess, and Lillicrap}]{barth2018distributed}
Barth-Maron, G.; Hoffman, M.~W.; Budden, D.; Dabney, W.; Horgan, D.; Tb, D.;
  Muldal, A.; Heess, N.; and Lillicrap, T. 2018.
\newblock Distributed distributional deterministic policy gradients.
\newblock \emph{arXiv preprint arXiv:1804.08617}.

\bibitem[{Bertsekas(2010)}]{bertsekas2010rollout}
Bertsekas, D.~P. 2010.
\newblock Rollout algorithms for discrete optimization: A survey.
\newblock \emph{Handbook of Combinatorial Optimization, D. Zu and P. Pardalos,
  Eds. Springer}.

\bibitem[{Clemente, Castej{\'o}n, and Chandra(2017)}]{clemente2017efficient}
Clemente, A.~V.; Castej{\'o}n, H.~N.; and Chandra, A. 2017.
\newblock Efficient parallel methods for deep reinforcement learning.
\newblock \emph{arXiv preprint arXiv:1705.04862}.

\bibitem[{De~Asis et~al.(2018)De~Asis, Hernandez-Garcia, Holland, and
  Sutton}]{de2018multi}
De~Asis, K.; Hernandez-Garcia, J.; Holland, G.; and Sutton, R. 2018.
\newblock Multi-step reinforcement learning: A unifying algorithm.
\newblock In \emph{Proceedings of the AAAI Conference on Artificial
  Intelligence}, volume~32.

\bibitem[{Duan et~al.(2016)Duan, Chen, Houthooft, Schulman, and
  Abbeel}]{duan2016benchmarking}
Duan, Y.; Chen, X.; Houthooft, R.; Schulman, J.; and Abbeel, P. 2016.
\newblock Benchmarking deep reinforcement learning for continuous control.
\newblock In \emph{International conference on machine learning}, 1329--1338.
  PMLR.

\bibitem[{Feinberg et~al.(2018)Feinberg, Wan, Stoica, Jordan, Gonzalez, and
  Levine}]{feinberg2018model}
Feinberg, V.; Wan, A.; Stoica, I.; Jordan, M.~I.; Gonzalez, J.~E.; and Levine,
  S. 2018.
\newblock Model-based value estimation for efficient model-free reinforcement
  learning.
\newblock \emph{arXiv preprint arXiv:1803.00101}.

\bibitem[{Fujimoto, Hoof, and Meger(2018)}]{fujimoto2018addressing}
Fujimoto, S.; Hoof, H.; and Meger, D. 2018.
\newblock Addressing function approximation error in actor-critic methods.
\newblock In \emph{International conference on machine learning}, 1587--1596.
  PMLR.

\bibitem[{Haarnoja et~al.(2018)Haarnoja, Zhou, Abbeel, and
  Levine}]{haarnoja2018soft}
Haarnoja, T.; Zhou, A.; Abbeel, P.; and Levine, S. 2018.
\newblock Soft actor-critic: Off-policy maximum entropy deep reinforcement
  learning with a stochastic actor.
\newblock In \emph{International conference on machine learning}, 1861--1870.
  PMLR.

\bibitem[{Henderson et~al.(2018)Henderson, Islam, Bachman, Pineau, Precup, and
  Meger}]{henderson2018deep}
Henderson, P.; Islam, R.; Bachman, P.; Pineau, J.; Precup, D.; and Meger, D.
  2018.
\newblock Deep reinforcement learning that matters.
\newblock In \emph{Proceedings of the AAAI conference on artificial
  intelligence}, volume~32.

\bibitem[{Hernandez-Garcia and Sutton(2019)}]{hernandez2019understanding}
Hernandez-Garcia, J.~F.; and Sutton, R.~S. 2019.
\newblock Understanding multi-step deep reinforcement learning: a systematic
  study of the DQN target.
\newblock \emph{arXiv preprint arXiv:1901.07510}.

\bibitem[{Hessel et~al.(2018)Hessel, Modayil, Van~Hasselt, Schaul, Ostrovski,
  Dabney, Horgan, Piot, Azar, and Silver}]{hessel2018rainbow}
Hessel, M.; Modayil, J.; Van~Hasselt, H.; Schaul, T.; Ostrovski, G.; Dabney,
  W.; Horgan, D.; Piot, B.; Azar, M.; and Silver, D. 2018.
\newblock Rainbow: Combining improvements in deep reinforcement learning.
\newblock In \emph{Thirty-second AAAI conference on artificial intelligence}.

\bibitem[{Lillicrap et~al.(2015)Lillicrap, Hunt, Pritzel, Heess, Erez, Tassa,
  Silver, and Wierstra}]{lillicrap2015continuous}
Lillicrap, T.~P.; Hunt, J.~J.; Pritzel, A.; Heess, N.; Erez, T.; Tassa, Y.;
  Silver, D.; and Wierstra, D. 2015.
\newblock Continuous control with deep reinforcement learning.
\newblock \emph{arXiv preprint arXiv:1509.02971}.

\bibitem[{Meng, Gorbet, and Kuli{\'c}(2021)}]{meng2021effect}
Meng, L.; Gorbet, R.; and Kuli{\'c}, D. 2021.
\newblock The effect of multi-step methods on overestimation in deep
  reinforcement learning.
\newblock In \emph{2020 25th International Conference on Pattern Recognition
  (ICPR)}, 347--353. IEEE.

\bibitem[{Mnih et~al.(2016)Mnih, Badia, Mirza, Graves, Lillicrap, Harley,
  Silver, and Kavukcuoglu}]{mnih2016asynchronous}
Mnih, V.; Badia, A.~P.; Mirza, M.; Graves, A.; Lillicrap, T.; Harley, T.;
  Silver, D.; and Kavukcuoglu, K. 2016.
\newblock Asynchronous methods for deep reinforcement learning.
\newblock In \emph{International conference on machine learning}, 1928--1937.
  PMLR.

\bibitem[{Mnih et~al.(2013)Mnih, Kavukcuoglu, Silver, Graves, Antonoglou,
  Wierstra, and Riedmiller}]{mnih2013playing}
Mnih, V.; Kavukcuoglu, K.; Silver, D.; Graves, A.; Antonoglou, I.; Wierstra,
  D.; and Riedmiller, M. 2013.
\newblock Playing atari with deep reinforcement learning.
\newblock \emph{arXiv preprint arXiv:1312.5602}.

\bibitem[{Pardo(2020)}]{pardo2020tonic}
Pardo, F. 2020.
\newblock Tonic: A deep reinforcement learning library for fast prototyping and
  benchmarking.
\newblock \emph{arXiv preprint arXiv:2011.07537}.

\bibitem[{Precup(2000)}]{precup2000eligibility}
Precup, D. 2000.
\newblock Eligibility traces for off-policy policy evaluation.
\newblock \emph{Computer Science Department Faculty Publication Series}, 80.

\bibitem[{Schulman et~al.(2015)Schulman, Moritz, Levine, Jordan, and
  Abbeel}]{schulman2015high}
Schulman, J.; Moritz, P.; Levine, S.; Jordan, M.; and Abbeel, P. 2015.
\newblock High-dimensional continuous control using generalized advantage
  estimation.
\newblock \emph{arXiv preprint arXiv:1506.02438}.

\bibitem[{Schulman et~al.(2017)Schulman, Wolski, Dhariwal, Radford, and
  Klimov}]{schulman2017proximal}
Schulman, J.; Wolski, F.; Dhariwal, P.; Radford, A.; and Klimov, O. 2017.
\newblock Proximal policy optimization algorithms.
\newblock \emph{arXiv preprint arXiv:1707.06347}.

\bibitem[{Silver et~al.(2017)Silver, Hasselt, Hessel, Schaul, Guez, Harley,
  Dulac-Arnold, Reichert, Rabinowitz, Barreto et~al.}]{silver2017predictron}
Silver, D.; Hasselt, H.; Hessel, M.; Schaul, T.; Guez, A.; Harley, T.;
  Dulac-Arnold, G.; Reichert, D.; Rabinowitz, N.; Barreto, A.; et~al. 2017.
\newblock The predictron: End-to-end learning and planning.
\newblock In \emph{International Conference on Machine Learning}, 3191--3199.
  PMLR.

\bibitem[{Silver et~al.(2018)Silver, Hubert, Schrittwieser, Antonoglou, Lai,
  Guez, Lanctot, Sifre, Kumaran, Graepel et~al.}]{silver2018general}
Silver, D.; Hubert, T.; Schrittwieser, J.; Antonoglou, I.; Lai, M.; Guez, A.;
  Lanctot, M.; Sifre, L.; Kumaran, D.; Graepel, T.; et~al. 2018.
\newblock A general reinforcement learning algorithm that masters chess, shogi,
  and Go through self-play.
\newblock \emph{Science}, 362(6419): 1140--1144.

\bibitem[{Sutton and Barto(2018)}]{sutton2018reinforcement}
Sutton, R.~S.; and Barto, A.~G. 2018.
\newblock \emph{Reinforcement learning: An introduction}.
\newblock MIT press.

\bibitem[{Tesauro(1994)}]{tesauro1994td}
Tesauro, G. 1994.
\newblock TD-Gammon, a self-teaching backgammon program, achieves master-level
  play.
\newblock \emph{Neural computation}, 6(2): 215--219.

\bibitem[{Van~Seijen and Sutton(2013)}]{van2013efficient}
Van~Seijen, H.; and Sutton, R.~S. 2013.
\newblock Efficient planning in MDPs by small backups.
\newblock In \emph{Proc. 30th Int. Conf. Mach. Learn}, volume~28. Citeseer.

\bibitem[{Watkins(1989)}]{watkins1989learning}
Watkins, C. J. C.~H. 1989.
\newblock Learning from delayed rewards.

\bibitem[{Yuan et~al.(2019)Yuan, Yu, Gu, Yeboah, Wei, Deng, Li, and
  Li}]{yuan2019novel}
Yuan, Y.; Yu, Z.~L.; Gu, Z.; Yeboah, Y.; Wei, W.; Deng, X.; Li, J.; and Li, Y.
  2019.
\newblock A novel multi-step Q-learning method to improve data efficiency for
  deep reinforcement learning.
\newblock \emph{Knowledge-Based Systems}, 175: 107--117.

\end{thebibliography}
\end{document}